\documentclass{amia}

\usepackage{booktabs}

\usepackage{multirow}

\graphicspath{{figures/}}

\usepackage{placeins}

\usepackage{booktabs}
\usepackage{longtable}
\usepackage{float} 
\usepackage{amsmath}   

\newcommand\blfootnote[1]{%
  \begingroup
    \renewcommand\thefootnote{}\footnote{#1}%
    \addtocounter{footnote}{-1}%
  \endgroup}
\makeatother

\begin{document}

\title{Patient-Reported Survey Data Improve Prediction of Opioid Use Disorder}

\author{Xiyue Jiang\amiasuper{1}\textdagger,
        Zihan Ding\amiasuper{2}\textdagger,
        Grace Han\amiasuper{3},
        Yinan Liu\amiasuper{2},
        Richard N. Rosenthal\amiasuper{4},
        Fusheng Wang, PhD\amiasuper{2,3\textdaggerdbl}} 

\institutes{%
  \amiasuper{1}Department of Applied Mathematics \& Statistics,
  Stony Brook University, Stony Brook, NY, USA;
  \amiasuper{2}Department of Computer Science,
  Stony Brook University, Stony Brook, NY, USA;
  \amiasuper{3}Department of Biomedical Informatics,
  Stony Brook University, Stony Brook, NY, USA;
  \amiasuper{4}Department of Psychiatry,
  Stony Brook Medicine, Stony Brook, NY, USA}

\maketitle
\blfootnote{\textsuperscript{\textdagger}Both authors contributed equally.} 
\blfootnote{\textsuperscript{\textdaggerdbl}Corresponding author: Fusheng Wang(fusheng.wang@stonybrook.edu).} 
\blfootnote{%
Analysis code is available at \url{https://github.com/StonyBrookDB/AllOfUsOUDPrediction}.
}
\blfootnote{This work was supported by the Patient-Centered Outcomes Research Institute (PCORI) under Contract No. ME-2023C3-35532.
} 


\section*{Abstract}

\textit{Electronic health records (EHRs) may incompletely capture patient-reported factors associated with opioid use disorder (OUD). We evaluated whether survey data improve prediction of a first recorded OUD diagnosis among 267,747 All of Us participants with documented opioid exposure, including 15,287 OUD cases. We compared EHR-only and EHR+survey models across 6-, 12-, and 24-month look-back windows using logistic regression, random forest, XGBoost, LightGBM, multilayer perceptron, LSTM, GRU, and Transformer. Survey augmentation improved PR-AUC across all 24 model-window combinations by 0.0087--0.0505; the best 24-month LightGBM model improved from 0.6219 to 0.6603. Survey coverage increased with longer windows and differed by OUD status (24 months: 21.7\% OUD-positive vs.\ 60.7\% OUD-negative). Permutation analysis ranked survey features as the second most important information domain at 24 months in both evaluated models. Patient-reported data provide complementary predictive signals beyond structured EHRs while highlighting the importance of survey availability.}

\section*{Introduction}

Opioid use disorder (OUD) is characterized by a problematic pattern of opioid
use that causes clinically significant impairment or distress\cite{cdc_oud_dx}.
It remains a substantial public health burden in the United States, affecting
an estimated 4.0 million people aged 12 years or older and resulting in 44,654
overdose deaths involving opioids in
2025\cite{nsduh2025,cdc_overdose_2025}. Identifying individuals at elevated
risk before OUD is first documented in the clinical record may create
opportunities for earlier clinical assessment and facilitate timely
intervention, thereby helping to reduce the risk of subsequent opioid-related
harms, including overdose and mortality.

Electronic health records (EHRs) provide rich longitudinal information (e.g.,
diagnoses, medications, laboratory results, procedures, physiological
measurements, and healthcare utilization) and have become an important data
source for machine learning (ML)-based prediction of OUD and other
opioid-related outcomes\cite{ramirezmedina2025}. Prior prediction studies have largely relied on routinely collected clinical and administrative data, highlighting patients' demographic characteristics, psychiatric and substance use histories, pain-related clinical history, medication exposure, and patterns of healthcare utilization as predictive features\cite{Song2024}. 
Our group has similarly used EHR histories to predict 
Opioid-related disease among patients prescribed opioids and, more recently, 
systematically evaluated how the selection of diagnosis-based features 
influences OUD prediction\cite{dong2019,dong2021,dong2021jbi,dong2023,ding2024,ding2026}. 
However, models based solely on routinely documented clinical data may capture only part of the risk profile associated with OUD.

OUD reflects the interaction of clinical, behavioral, and social factors. Among patients prescribed opioids, prior substance use disorders and psychiatric comorbidities (e.g., depression, anxiety disorders, and personality disorders) have been associated with subsequent OUD or problematic opioid use \cite{Klimas2019,vanRijswijk2019}. Beyond these clinical and behavioral characteristics, social conditions may further shape vulnerability to opioid-related harms. A recent umbrella review found consistent associations between OUD and broader social
determinants of health (SDOH), including unemployment and adverse childhood
experiences, with additional socioeconomic factors linked to opioid
overdose\cite{loeffel2026}. Yet such factors are often incompletely or
inconsistently represented in EHR data, which can introduce missingness and
misclassification when these records are used for research\cite{cook2021}.
As a result, EHRs may not fully reflect the broader context associated with OUD risk.

Patient-reported information can complement EHRs by capturing additional factors that may be underrepresented in clinical data. The \textit{All of Us} Research Program is well-suited to evaluate their added value because it combines longitudinal EHR data with participant-reported
surveys in a large and diverse cohort\cite{ramirez2022,cronin2019}. Participants complete structured questionnaires through the program’s online portal, providing information on health behaviors, substance use, socioeconomic and social conditions, physical and mental health, daily functioning, disability, and healthcare access that may not be routinely documented in the EHR. Prior work
has shown that patient-reported information derived from preoperative
questionnaires (e.g., measures of mental health, pain, substance use, and
socioeconomic circumstances) can improve EHR-based prediction of persistent
opioid use after surgery\cite{giladi2023}. Persistent opioid use, however, is
distinct from OUD, and whether survey information provides incremental
predictive value for OUD beyond longitudinal EHR data remains unknown.

In this study, we evaluated whether incorporating participant-reported survey
information improves prediction of a first recorded OUD diagnosis beyond EHR
data alone. We compared EHR-only and EHR+survey models across multiple machine
learning (ML) approaches using 6-, 12-, and 24-month look-back windows,
selected to capture progressively longer pre-diagnostic histories. These windows were chosen because prior
claims-based research found that OUD diagnosis occurred, on average,
approximately 10--12 months after initial prescription opioid exposure among
patients who subsequently developed OUD\cite{schoenfeld2026}. Our primary objective was to determine whether
patient-reported information consistently improves OUD prediction beyond
information already captured in the EHR.

Our study makes three contributions. First, it quantifies the incremental predictive value of participant-reported information beyond longitudinal clinical and demographic features. Second, it tests whether this value is consistent across linear, tree-based, neural-network, and sequential models and across multiple observation windows. Third, it examines the information domains and individual survey questions that contribute most strongly to OUD prediction.


\section*{Methods}

\subsection*{\textit{Data source and study cohort}}

Data were obtained from the \textit{All of Us} Research Program Curated Data Repository,
with data available through October 1, 2023. The study cohort
comprised 267,747 participants with at least one documented exposure to an
opioid medication, identified using the Anatomical Therapeutic Chemical (ATC)
level 3 code N02A. Of those, 15,287 participants had a recorded OUD diagnosis
(based on ICD-9-CM codes 304.00--304.03 or ICD-10-CM F11 code family) and were classified as
OUD-positive; the remaining 252,460 were classified as OUD-negative, yielding a
case-to-control ratio of approximately 1:16.5. Baseline characteristics of the cohort are summarized in Table~\ref{tab:cohort}.
\begin{longtable}{@{}l r r r@{}}
    \caption{Baseline characteristics of the cohort, overall and by opioid use
    disorder (OUD) status.} 
    \label{tab:cohort} \\
    \toprule
    \textbf{Variable} & \textbf{Overall} & \textbf{OUD-positive} & \textbf{OUD-negative} \\
    \midrule
    \endfirsthead
    
    \multicolumn{4}{c}{{\tablename\ \thetable{} -- continued from previous page}} \\
    \toprule
    \textbf{Variable} & \textbf{Overall} & \textbf{OUD-positive} & \textbf{OUD-negative} \\
    \midrule
    \endhead
    
    \midrule
    \multicolumn{4}{r}{\textit{Continued on next page}} \\
    \endfoot
    
    \bottomrule
    \endlastfoot
    
    Number of patients & 267,747 & 15,287 & 252,460 \\
    \addlinespace
    \textbf{Gender} & & & \\
    \hspace{1em}Female & 165,095 (61.7\%) & 7,562 (49.5\%) & 157,533 (62.4\%) \\
    \hspace{1em}Male & 97,451 (36.4\%) & 7,321 (47.9\%) & 90,130 (35.7\%) \\
    \hspace{1em}Others/Unknown & 5,201 (1.9\%) & 404 (2.6\%) & 4,797 (1.9\%) \\
    \addlinespace
    \textbf{Race} & & & \\
    \hspace{1em}Asian & 5,329 (2.0\%) & 73 (0.5\%) & 5,256 (2.1\%) \\
    \hspace{1em}Black or African American & 46,156 (17.2\%) & 3,696 (24.2\%) & 42,460 (16.8\%) \\
    \hspace{1em}White & 150,777 (56.3\%) & 7,449 (48.7\%) & 143,328 (56.8\%) \\
    \hspace{1em}More than one race & 11,700 (4.4\%) & 936 (6.1\%) & 10,764 (4.3\%) \\
    \hspace{1em}Others/Unknown & 49,531 (18.5\%) & 2,618 (17.1\%) & 46,913 (18.6\%) \\
    \addlinespace
    Prior opioid overdose & 1,199 (0.4\%) & 1,044 (6.8\%) & 155 (0.1\%) \\
    \addlinespace
    \textbf{Any other substance use disorder} & 25,226 (9.4\%) & 6,809 (44.5\%) & 18,417 (7.3\%) \\
    \hspace{1em}Alcohol use disorder & 19,807 (7.4\%) & 4,513 (29.5\%) & 15,294 (6.1\%) \\
    \hspace{1em}Cocaine use disorder & 8,147 (3.0\%) & 3,517 (23.0\%) & 4,630 (1.8\%) \\
    \addlinespace
    \textbf{Any mental health condition} & 110,990 (41.5\%) & 11,588 (75.8\%) & 99,402 (39.4\%) \\
    \hspace{1em}Depression & 83,287 (31.1\%) & 9,578 (62.7\%) & 73,709 (29.2\%) \\
    \hspace{1em}Bipolar disorder & 14,888 (5.6\%) & 3,471 (22.7\%) & 11,417 (4.5\%) \\
    \hspace{1em}Anxiety disorder & 81,678 (30.5\%) & 9,149 (59.8\%) & 72,529 (28.7\%) \\
    \hspace{1em}PTSD & 4,532 (1.7\%) & 822 (5.4\%) & 3,710 (1.5\%) \\
    \addlinespace
    \textbf{Any pain diagnosis} & 136,318 (50.9\%) & 11,363 (74.3\%) & 124,955 (49.5\%) \\
    \hspace{1em}Back pain & 75,319 (28.1\%) & 7,106 (46.5\%) & 68,213 (27.0\%) \\
    \hspace{1em}Chronic pain (general) & 89,665 (33.5\%) & 9,623 (62.9\%) & 80,042 (31.7\%) \\
    \hspace{1em}Muscle/musculoskeletal/joint pain & 36,063 (13.5\%) & 3,784 (24.8\%) & 32,279 (12.8\%) \\
    \addlinespace
    Benzodiazepine use & 31,046 (11.6\%) & 3,645 (23.8\%) & 27,401 (10.9\%) \\
    Pain score, mean (SD) & 3.67 (3.02) & 5.80 (2.85) & 3.54 (2.98) \\
    Insured (self-reported) & 248,889 (95.3\%) & 13,772 (93.8\%) & 235,117 (95.4\%) \\
    \addlinespace
    \textbf{Income} & & & \\
    \hspace{1em}\textless 25k & 66,900 (31.4\%) & 8,068 (69.4\%) & 58,832 (29.2\%) \\
    \hspace{1em}25k--50k & 40,530 (19.0\%) & 1,850 (15.9\%) & 38,680 (19.2\%) \\
    \hspace{1em}50k--100k & 50,089 (23.5\%) & 1,160 (10.0\%) & 48,929 (24.3\%) \\
    \hspace{1em}\textgreater 100k & 55,374 (26.0\%) & 542 (4.7\%) & 54,832 (27.2\%) \\
    \addlinespace
    Stable housing concern & 44,630 (16.9\%) & 6,337 (42.5\%) & 38,293 (15.4\%) \\
    
    \end{longtable}

\subsection*{\textit{Temporal design}}

\begin{figure}[H]
  \centering
  \includegraphics[width=0.75\textwidth]{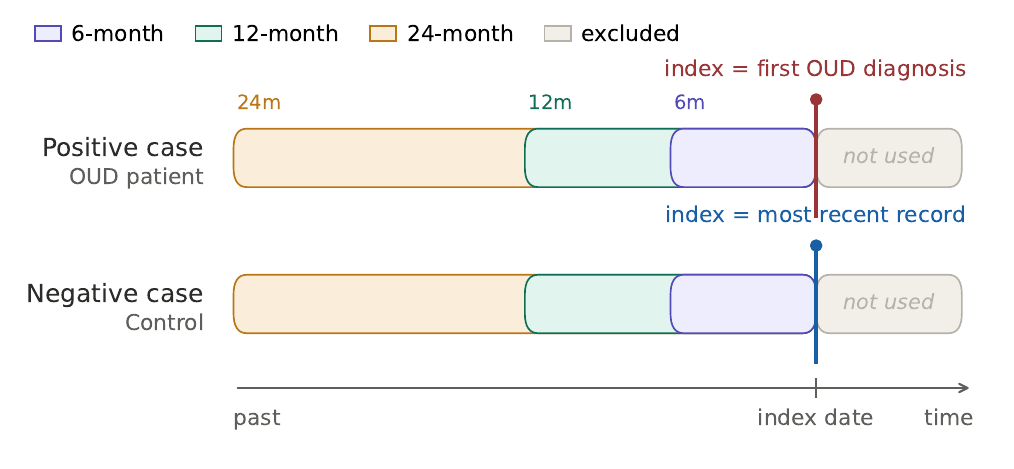}
  \caption{Temporal design for OUD-positive participants and OUD-negative
           controls across 6-, 12-, and 24-month look-back windows.}
  \label{fig:temporal}
\end{figure}

An index date was defined for each participant to anchor longitudinal feature
extraction. For OUD-positive cases, the index date was the first recorded OUD
diagnosis; for OUD-negative controls, it was the most recent medical encounter
available in the record. Features were extracted from the 6, 12, and 24 months preceding the index date (Figure~\ref{fig:temporal}), representing progressively longer pre-diagnostic histories. These windows were informed in part by unpublished analyses from our group using the Health Facts database, in which 34.12\% of patients who later received an OUD diagnosis were diagnosed within 1 year and 54.07\% within 2 years of their first documented opioid medication. Data recorded on or after
the index date were excluded to prevent temporal leakage.


\subsection*{\textit{EHR-Derived Features}}

The analysis incorporated EHR-derived features (i.e., demographics, diagnoses, medications, laboratory measurements, physical measurements, and clinical observations). Candidate concepts within each domain were selected \textit{a priori} based on their clinical relevance to OUD. All features were constructed from information recorded within the specified look-back window before the index date. To reduce sparsity, clinical concepts observed in fewer than 0.5\% of participants within each look-back window were excluded (Table~\ref{tab:features}). Feature-inclusion thresholds were calculated using the full cohort before the train/validation/test split and were based only on feature frequency, without using OUD status.


\begin{table}[t]
  \caption{Candidate and retained concepts/questions by data source across the 6-, 12-,
           and 24-month look-back windows.}
  \label{tab:features}
  \begin{center}
  \small
  \begin{tabular}{lccc}
  \toprule
  \textbf{Source} & \textbf{Candidate (6/12/24 mo)}
    & \textbf{Retained (6/12/24 mo)} & \textbf{Retained, \% (6/12/24 mo)} \\
  \midrule
  Demographics & 6 / 6 / 6          & 6 / 6 / 6       & 100 / 100 / 100    \\
  Conditions   & 911 / 987 / 1101   & 48 / 72 / 84    & 5.3 / 7.3 / 7.6    \\
  Medications  & 4571 / 5020 / 5421 & 156 / 229 / 320 & 3.4 / 4.6 / 5.9    \\
  Lab values   & 134 / 135 / 135    & 60 / 60 / 62    & 44.8 / 44.4 / 45.9 \\
  Measurements & 9 / 9 / 9          & 9 / 9 / 9       & 100 / 100 / 100    \\
  Observations & 16 / 16 / 16       & 9 / 10 / 13     & 56.3 / 62.5 / 81.3 \\
  Surveys      & 188 / 188 / 188    & 30 / 30 / 108   & 16.0 / 16.0 / 57.4 \\
  \bottomrule
  \end{tabular}
  \end{center}
\end{table}

\textit{Demographics}. Demographic features included age, gender, race, ethnicity, sex at birth, and self-reported
category (a composite race/ethnicity variable).

\textit{Conditions}. Condition features included diagnoses such as
low back pain, sciatica, fibromyalgia, chronic pain syndrome, alcohol abuse,
psychoactive substance abuse, major depressive disorder, generalized anxiety
disorder, and posttraumatic stress disorder. These features were represented
by their occurrence counts within each look-back window. Participants with no
recorded occurrence of a given concept were assigned a count of zero.

\textit{Medications}. Medication features included opioid
analgesics (e.g., oxycodone, hydromorphone, morphine, fentanyl, and tramadol),
non-opioid analgesics (e.g., ibuprofen, ketorolac, and celecoxib), muscle
relaxants (e.g., cyclobenzaprine, tizanidine), benzodiazepines (e.g.,
lorazepam, diazepam), and antidepressants (e.g., sertraline, duloxetine).
These features were represented using the same occurrence-count approach
described above.

\textit{Laboratory and physical measurements}. Laboratory features included standard
metabolic, liver, lipid, and hematology panels (e.g., creatinine, alanine
aminotransferase, total cholesterol, and complete blood count), as well as a
urine toxicology panel (e.g., opiates, fentanyl, oxycodone, methadone,
buprenorphine, cocaine, cannabinoids, benzodiazepines, and amphetamines).
Physical measurement features
included body weight, height, body mass index, systolic and diastolic blood
pressure, heart rate, and waist and hip circumference. Any values recorded in different units were first converted to a
common unit and then summarized within each look-back window using three
features: count, mean value, and most recent value. When no measurement was
available, the count was set to zero, while the mean and most recent values
were retained as missing.

\textit{Clinical observations}. Clinical observation features
included both numeric and occurrence-based concepts. Numeric observations
(e.g., pain score, tobacco smoking, depression screening assessment)
were summarized using count, mean, and most recent value, following the same
approach as laboratory and physical measurement features. Occurrence-based
observations (e.g., housing instability, documented abuse, treatment
noncompliance, and self-reported alcohol use) were represented using occurrence
counts within each look-back window, with a count of zero assigned when no
recorded occurrence was present.

\subsection*{\textit{Participant-reported survey data}}

\textit{Survey Sources and Content}. 
Participant-reported information was obtained from four \textit{All of Us} Research Program surveys: \textit{The Basics}, \textit{Lifestyle}, \textit{Overall Health}, and \textit{Social Determinants of Health}\textsuperscript{16,17}. \textit{The Basics} is administered early in participation and captures demographic and socioeconomic characteristics, including employment, insurance, housing, and home-life information. \textit{Lifestyle} and \textit{Overall Health} become available after completion of \textit{The Basics}. \textit{Lifestyle} assesses health behaviors including tobacco, alcohol, and recreational drug use, whereas \textit{Overall Health} assesses general health, daily functioning, pain, and physical and mental health. \textit{Social Determinants of Health} is a follow-up survey available after completion of the baseline surveys and captures broader social and environmental factors, including neighborhood characteristics, social relationships, stress, discrimination, loneliness, and social support. 

\textit{Survey Availability}. 
Only survey responses recorded before the index date and within the corresponding 6-, 12-, or 24-month look-back window were eligible for feature construction. Because the source surveys differ in administration sequence, eligibility, and timing of completion, survey information was not uniformly available across participants or prediction windows. We therefore characterized both cohort-level survey coverage and respondent-level completion behavior. Survey coverage was defined as the proportion of participants with at least one eligible survey interaction within the corresponding look-back window, including either an answered item or an explicitly recorded skip or decline response. Among these respondents, we then summarized the number of questions answered, the number of questions explicitly skipped or declined, and the number of days between the most recent eligible survey response and the index date. These measures separately characterize whether survey data were present, how much information was available, and how recently it was collected.

\textit{Survey-Derived Features}. 
Because the objective was to determine whether participant-reported information provides predictive value beyond routinely documented EHR-derived features, eligible survey questions were retained even when they reflected constructs also represented in the EHR, such as pain or substance use. Among participants who contributed any survey data within a given look-back window (individuals with at least one valid survey response in that window), questions with below 50\% of response rates were excluded. The remaining questions were encoded using binary indicators for individual response options. Single-response questions were represented using one-hot encoding, whereas multiple-response questions were represented using multi-hot encoding. A value of 1 indicated that a participant selected the corresponding response option, 0 indicated that the question was answered but the option was not selected, and missing indicated that no usable response was available for that item within the look-back window. The number of retained survey features consequently varied by observation window, with 30, 30, and 108 retained survey features at 6, 12, and 24 months, respectively (Table~\ref{tab:features}).

\subsection*{\textit{Model development}}

We evaluated two complementary modeling approaches. Tabular models used patient-level features summarized over each look-back window, whereas sequential models incorporated the temporal ordering of clinical events to assess whether longitudinal information improved OUD prediction.

\textit{Cohort splitting and preprocessing.}
Across all look-back windows, participants were stratified by OUD status and split at the patient level into training (68\%), validation (12\%), and independent test (20\%) sets. The same patient splits were used for the corresponding EHR-only and EHR+survey models to enable paired comparisons. The validation set was used for early stopping and hyperparameter selection, where applicable, as well as for decision-threshold selection. All preprocessing parameters, including those used for imputation and standardization, were estimated from the training set only and then applied unchanged to the validation and test sets.

\textit{Class imbalance.}
We evaluated four approaches to class imbalance: no adjustment, cost-sensitive class weighting, synthetic minority oversampling using the synthetic minority over-sampling technique (SMOTE), and random undersampling. The comparison was performed using XGBoost with EHR and survey features at the 24-month look-back window. Strategies were compared using validation area under the precision-recall curve (PR-AUC). Unadjusted training achieved the highest validation PR-AUC (0.6533), followed by class weighting (0.6475), random undersampling (0.5906), and SMOTE (0.4075); therefore, final models were trained without class-imbalance adjustment.

\textit{Tabular models.}
Five tabular models were evaluated: logistic regression, random forest, XGBoost, LightGBM, and a multilayer perceptron (MLP). Logistic regression served as the linear baseline. Missing EHR numeric features were median-imputed and standardized for logistic regression and the MLP, whereas random forest inputs were median-imputed without standardization. Missing survey feature values were also median-imputed. XGBoost, LightGBM, and the MLP used early stopping based on validation PR-AUC. Logistic regression was trained using the L-BFGS solver with L2 penalty and \(C=1.0\). The random forest consisted of 400 trees grown without a maximum depth. XGBoost and LightGBM were each trained for up to 5,000 boosting rounds with early stopping after 200 rounds without improvement in validation PR-AUC. Both used a learning rate of 0.03 and row and column subsampling rates of 0.75 and 0.70, respectively. XGBoost additionally used a maximum depth of 5 and a minimum child weight of 16, whereas LightGBM used 31 leaves per tree. The MLP consisted of two hidden layers with 256 and 128 units, and a dropout rate of 0.3. It was trained using the Adam optimizer and binary cross-entropy loss with a batch size of 2,048 for up to 60 epochs, with early stopping based on validation PR-AUC.

\textit{Sequential models.}
Three sequential architectures were evaluated: long short-term memory (LSTM), gated recurrent unit (GRU), and Transformer models. For each participant, the longitudinal input consisted of the \(K\) most recent dates with recorded clinical activity within the corresponding look-back window. Sequence lengths were set to \(K=20\), 30, and 50 for the 6-, 12-, and 24-month windows, respectively. These thresholds were chosen so that approximately 90\% of participants had no more than \(K\) recorded activity dates within each window. This resulted in coverage of 93.2\%, 90.9\%, and 89.7\%, respectively. Participants with fewer than \(K\) dates were zero-padded, while those with more than \(K\) dates were truncated to the \(K\) most recent dates. Padded positions were masked during model training.

Each time step represented clinical events recorded on a single date. The input for each date was a vector of clinical concept counts, with each feature corresponding to a diagnosis, medication, laboratory test, physical measurement, or clinical observation code and its value indicating the number of times that code was recorded that day. Laboratory and physical measurement codes captured the occurrence of these events, while their numeric values were summarized across the full look-back window using the mean and most recent observed value and included in the static feature vector along with demographic characteristics. For EHR+survey models, the same survey-derived features used in the tabular models were added to the static feature vector.

The LSTM and GRU processed the sequence chronologically and used the final hidden state as the patient-level temporal representation. The Transformer projected each time-step vector to a 128-dimensional embedding, added learnable positional embeddings, and processed the sequence using a two-layer Transformer encoder. Transformer outputs were aggregated into a single patient-level temporal representation using masked mean pooling over non-padded time steps. For all three architectures, the temporal representation was concatenated with an encoded representation of the static features, obtained via a fully-connected layer, before the final prediction layer. EHR-only and EHR+survey models used the same longitudinal inputs and architecture, differing only in the addition of survey-derived features to the static vector.

The three sequential models shared a common overall architecture, differing primarily in their temporal encoders. The LSTM and GRU used a hidden dimension of 128, while the Transformer used a 128-dimensional embedding with four attention heads and two encoder layers. In all three models, the temporal representation was concatenated with a 64-dimensional static representation produced by a fully connected layer with ReLU activation and a dropout rate of 0.3. The combined representation was then passed through a two-layer prediction head. All models were trained using the Adam optimizer with a learning rate of \(1\times10^{-3}\) and weight decay of \(1\times10^{-5}\), using binary cross-entropy loss and a batch size of 512. Training was performed for up to 30 epochs, with early stopping after five epochs without improvement in validation PR-AUC.

\subsection*{\textit{Evaluation}}

Model performance was evaluated on the held-out test set using PR-AUC, area under the receiver operating characteristic curve (ROC-AUC), positive-class precision, recall, and F1-score. PR-AUC was considered the primary performance metric due to class imbalance. For each model, the classification threshold was selected on the validation set to maximize positive-class F1 across 181 thresholds from 0.05 to 0.95. The selected threshold was then applied unchanged to the test set. F1+ denotes the F1 score for the positive OUD class.

Feature importance was evaluated at the 6-, 12-, and 24-month look-back windows at two levels: broad information domains and individual survey questions. Both analyses used permutation importance on the held-out test set and were performed for the LightGBM and GRU models across all look-back windows. Within each look-back window, only the feature or feature group being evaluated was permuted across participants, while all other features remained unchanged. Importance was defined as the mean decrease in PR-AUC after permutation, averaged over five repetitions. For the domain-level analysis, features were grouped into demographics, conditions, medications, laboratory measurements, physical measurements, clinical observations, and survey-derived features. All features within a domain were jointly permuted to estimate the model's reliance on that source of information. For the survey question-level analysis, each survey question was evaluated separately. When a question was represented by multiple encoded response categories, all corresponding columns were permuted together as a single unit. This analysis identified the individual survey questions that contributed most to model performance at each look-back window. The same permutation framework was applied across tabular and sequential models. For sequential models, time-varying EHR modalities were permuted at the patient level by reassigning entire temporal trajectories rather than individual time points, thereby preserving within-patient temporal structure.

\section*{Results}

\subsection*{Survey Availability}

Survey coverage increased with longer look-back windows in both groups but was consistently lower among OUD-positive participants (Table~\ref{tab:survey_availability}). The proportion of participants with at least one eligible survey response increased from 9.5\% at 6 months to 21.7\% at 24 months among OUD-positive participants, compared with 21.6\% to 60.7\% among OUD-negative participants. Among respondents, the median number of questions answered was similar between groups at the 6- and 12-month windows, whereas at 24 months OUD-negative participants contributed a larger volume of survey responses (median 77 vs.\ 28 questions). Explicitly skipped or declined questions were uncommon across all windows, with median values of 0--1 per respondent. The median interval between the most recent eligible survey response and the index date also increased with longer look-back windows, from 51 and 63 days at 6 months to 217 and 270 days at 24 months for OUD-positive and OUD-negative participants, respectively.

\begin{table*}[!t]
\centering
\caption{Survey coverage and completion behavior by OUD status and look-back window.}
\label{tab:survey_availability}
\small
\setlength{\tabcolsep}{6pt}
\begin{tabular}{llrrrr}
\toprule
\textbf{Look-back} & \textbf{Group} &
\multicolumn{1}{c}{\textbf{Survey coverage}} &
\multicolumn{3}{c}{\textbf{Among respondents, median (IQR)}} \\
\cmidrule(lr){3-3}
\cmidrule(lr){4-6}
& &
\textbf{$\geq$1 eligible response, n (\%)} &
\textbf{Questions answered} &
\textbf{Skipped/declined} &
\textbf{Days to index} \\
\midrule
6 months
& OUD-positive & 1{,}456 (9.5\%)   & 28 (26--28)  & 1 (0--2) & 51 (15--110) \\
& OUD-negative & 54{,}455 (21.6\%) & 29 (26--30)  & 0 (0--1) & 63 (14--112) \\
\addlinespace

12 months
& OUD-positive & 2{,}230 (14.6\%)  & 28 (26--28)  & 1 (0--2) & 114 (31--228) \\
& OUD-negative & 93{,}109 (36.9\%) & 29 (2--30)   & 0 (0--1) & 141 (46--264) \\
\addlinespace

24 months
& OUD-positive & 3{,}324 (21.7\%)   & 28 (26--29)  & 1 (0--2) & 217 (65--426) \\
& OUD-negative & 153{,}277 (60.7\%) & 77 (29--105) & 1 (0--2) & 270 (94--466) \\
\bottomrule
\end{tabular}

\vspace{3pt}
\begin{minipage}{0.97\textwidth}
\footnotesize
\textit{Note.} Survey coverage was calculated among all OUD-positive
($n=15{,}287$) and OUD-negative ($n=252{,}460$) participants at each
look-back window. 
\end{minipage}
\end{table*}

\subsection*{Model Evaluation}

Across the evaluated models, ROC-AUC ranged from 0.8180 to 0.9430, while PR-AUC 
ranged from 0.3239 to 0.6603, reflecting the pronounced class imbalance in the study 
cohort. Among the eight models, LightGBM and XGBoost achieved the strongest overall 
performance. At the 24-month look-back window, LightGBM achieved the highest PR-AUC 
of 0.6603 with survey features, followed by XGBoost at 0.6576. Across all models, PR-AUC 
increased with longer look-back windows (Table~\ref{tab:performance}).

\begin{table}[H]
  \centering
  \caption{Predictive performance by model, feature set, and look-back window
           (months).}
  \label{tab:performance}
  \small
  \setlength{\tabcolsep}{5.5pt}
  \begin{tabular}{llccccccccc}
  \toprule
  & & \multicolumn{3}{c}{\textbf{PR-AUC}}
    & \multicolumn{3}{c}{\textbf{ROC-AUC}}
    & \multicolumn{3}{c}{\textbf{F1+}} \\
  \cmidrule(lr){3-5}
  \cmidrule(lr){6-8}
  \cmidrule(lr){9-11}

  \textbf{Model} & \textbf{Features}
  & \textbf{6m} & \textbf{12m} & \textbf{24m}
  & \textbf{6m} & \textbf{12m} & \textbf{24m}
  & \textbf{6m} & \textbf{12m} & \textbf{24m} \\
  \midrule

  \multirow{2}{*}{LogisticReg}
    & EHR   & 0.3239 & 0.3503 & 0.3672 & 0.8180 & 0.8343 & 0.8438 & 0.3552 & 0.3812 & 0.3910 \\
    & EHR+S & 0.3379 & 0.3769 & 0.4176 & 0.8355 & 0.8590 & 0.8848 & 0.3708 & 0.4076 & 0.4418 \\
  \addlinespace

  \multirow{2}{*}{RandomForest}
    & EHR   & 0.4599 & 0.5055 & 0.5416 & 0.8643 & 0.8855 & 0.9024 & 0.4405 & 0.4694 & 0.5024 \\
    & EHR+S & 0.4719 & 0.5269 & 0.5851 & 0.8773 & 0.9023 & 0.9270 & 0.4445 & 0.4899 & 0.5377 \\
  \addlinespace

  \multirow{2}{*}{XGBoost}
    & EHR   & 0.5277 & 0.5794 & 0.6226 & 0.8904 & 0.9091 & 0.9259 & 0.4956 & 0.5361 & 0.5649 \\
    & EHR+S & 0.5396 & 0.6010 & 0.6576 & 0.8997 & 0.9222 & 0.9422 & 0.5038 & 0.5556 & 0.5892 \\
  \addlinespace

  \multirow{2}{*}{LightGBM}
    & EHR   & 0.5277 & 0.5860 & 0.6219 & 0.8908 & 0.9117 & 0.9277 & 0.4926 & 0.5393 & 0.5685 \\
    & EHR+S & 0.5429 & 0.5998 & 0.6603 & 0.9014 & 0.9224 & 0.9430 & 0.5055 & 0.5522 & 0.5913 \\
  \addlinespace

  \multirow{2}{*}{MLP}
    & EHR   & 0.4375 & 0.4909 & 0.5106 & 0.8566 & 0.8781 & 0.8832 & 0.4365 & 0.4669 & 0.4917 \\
    & EHR+S & 0.4466 & 0.4996 & 0.5291 & 0.8579 & 0.8838 & 0.9061 & 0.4330 & 0.4854 & 0.5011 \\
  \addlinespace

  \multirow{2}{*}{LSTM}
    & EHR   & 0.4653 & 0.5151 & 0.5683 & 0.8700 & 0.8903 & 0.9074 & 0.4416 & 0.4829 & 0.5276 \\
    & EHR+S & 0.4741 & 0.5246 & 0.5863 & 0.8719 & 0.8932 & 0.9228 & 0.4521 & 0.5023 & 0.5427 \\
  \addlinespace

  \multirow{2}{*}{GRU}
    & EHR   & 0.4657 & 0.5153 & 0.5583 & 0.8720 & 0.8888 & 0.9053 & 0.4522 & 0.4900 & 0.5164 \\
    & EHR+S & 0.4767 & 0.5249 & 0.5967 & 0.8765 & 0.8915 & 0.9245 & 0.4590 & 0.4952 & 0.5483 \\
  \addlinespace

  \multirow{2}{*}{Transformer}
    & EHR   & 0.4654 & 0.5183 & 0.5587 & 0.8715 & 0.8936 & 0.9102 & 0.4506 & 0.4866 & 0.5087 \\
    & EHR+S & 0.4835 & 0.5297 & 0.5914 & 0.8768 & 0.9017 & 0.9257 & 0.4577 & 0.4947 & 0.5453 \\
  \bottomrule

  \end{tabular}
\end{table}

Adding survey-derived features improved PR-AUC in all 24 model--window combinations, 
with absolute gains ranging from 0.0087 to 0.0505 (Table~\ref{tab:gains}). Similar 
improvements were observed in ROC-AUC across linear, ensemble, gradient-boosting, 
and neural-network models, with gains ranging from 0.0013 to 0.0410. The incremental 
contribution of survey information was greatest at the 24-month look-back window, 
where all eight models achieved their largest PR-AUC improvement, with gains ranging 
from 0.0181 to 0.0505. However, the magnitude of improvement did not increase 
monotonically across look-back windows for all models.

%
%
%
%

\begin{table}[!htbp]
  \caption{Absolute increase in PR-AUC and ROC-AUC after adding survey-derived
           features, by model and look-back window (months). In each column,
           \textbf{bold} marks the largest gain and \underline{underline} the
           second largest.}
  \label{tab:gains}
  \begin{center}
  \small
  \begin{tabular}{lcccccc}
  \toprule
  & \multicolumn{3}{c}{\textbf{PR-AUC gain}} & \multicolumn{3}{c}{\textbf{ROC-AUC gain}} \\
  \cmidrule(lr){2-4}\cmidrule(lr){5-7}
  \textbf{Model} & \textbf{6m} & \textbf{12m} & \textbf{24m} & \textbf{6m} & \textbf{12m} & \textbf{24m} \\
  \midrule
  LogisticReg  & +0.0140 & \textbf{+0.0266} & \textbf{+0.0505} & \textbf{+0.0175} & \textbf{+0.0247} & \textbf{+0.0410} \\
  RandomForest & +0.0120 & +0.0214 & \underline{+0.0436} & \underline{+0.0130} & \underline{+0.0168} & \underline{+0.0246} \\
  XGBoost      & +0.0119 & \underline{+0.0216} & +0.0350 & +0.0093 & +0.0131 & +0.0163 \\
  LightGBM     & \underline{+0.0153} & +0.0138 & +0.0384 & +0.0106 & +0.0107 & +0.0153 \\
  MLP          & +0.0091 & +0.0087 & +0.0185 & +0.0013 & +0.0056 & +0.0229 \\
  LSTM         & +0.0088 & +0.0095 & +0.0181 & +0.0019 & +0.0028 & +0.0154 \\
  GRU          & +0.0110 & +0.0095 & +0.0384 & +0.0045 & +0.0027 & +0.0192 \\
  Transformer  & \textbf{+0.0181} & +0.0114 & +0.0326 & +0.0053 & +0.0081 & +0.0155 \\
  \bottomrule
  \end{tabular}
  \end{center}
\end{table}

Model complexity did not consistently translate into better predictive
performance. Despite their ability to model nonlinear relationships or temporal
structure, the LSTM, GRU, and Transformer models generally performed below the
gradient-boosting models, particularly at the longer look-back windows.
Nevertheless, survey-derived features improved performance even for the neural
and sequential models, suggesting that their contribution was complementary to
rather than dependent on a specific modeling architecture.

Domain-level permutation analysis showed that the contribution of survey-derived features increased with longer look-back windows. In LightGBM, permutation of survey features resulted in PR-AUC decreases of 0.084, 0.156, and 0.172 at the 6-, 12-, and 24-month windows, respectively. The corresponding decreases in the GRU were 0.046, 0.073, and 0.234, with the largest decrease observed at 24 months. In both models, survey features ranked as the second most important information domain at 24 months, behind laboratory measurements.

\begin{figure}[H]
    \centering
    \includegraphics[width=0.75\textwidth]{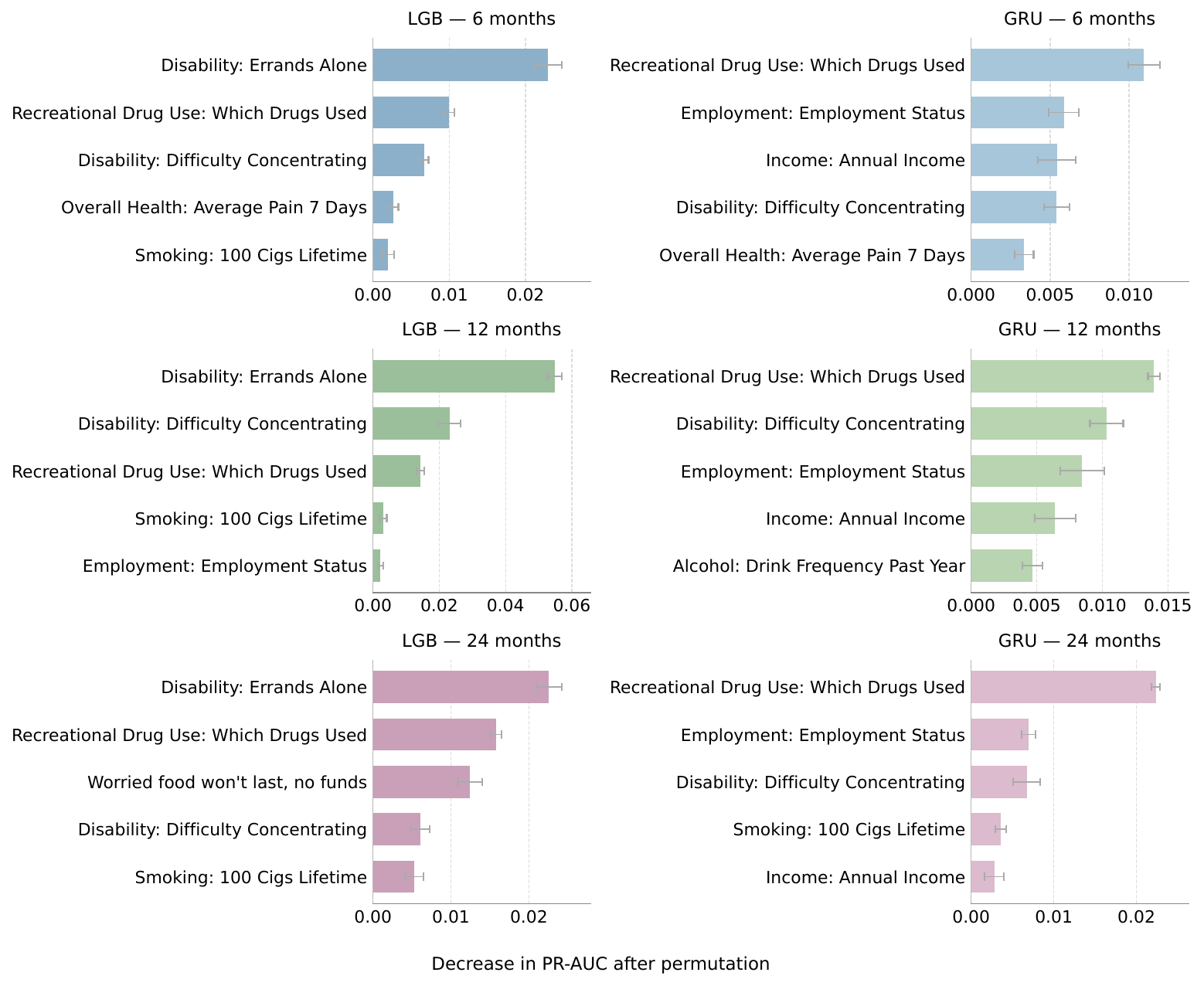}
    \caption{Permutation feature importance for the top 5 predictors of each model — LGB (left) and GRU (right) — at 6, 12, and 24 months. Bars show mean decrease in PR-AUC across five permutations; error bars indicate SD.}
    \label{fig:feature_importance}
\end{figure}

At the individual question level, the most influential survey features varied across models and look-back windows, although several questions consistently ranked among the top contributors (Figure~\ref{fig:feature_importance}). Recreational drug use, difficulty concentrating, ability to complete errands independently, and employment status repeatedly appeared among the highest-ranking questions. Other recurring features included measures of alcohol and tobacco use, pain and general health, income, and education. Overall, the highest-ranking survey questions spanned behavioral, functional, socioeconomic, and general health domains.


\section*{Discussion}

This study examined whether participant-reported survey information adds predictive value beyond structured EHR data for identifying patients at risk of a first recorded OUD diagnosis. Survey augmentation improved discrimination across all modeling approaches and observation windows, indicating that the added signal was not specific to one algorithmic family. Although the gains were modest, their consistency supports patient-reported information as a useful complement to EHR-based OUD prediction.

Feature-importance analyses further clarified the source of this added value. At the 24-month window, survey-derived features ranked among the most important information sources in both evaluated model architectures. Influential questions spanned substance use, functional limitations, employment, income, pain, and general health, suggesting that the predictive signal was distributed across behavioral, functional, socioeconomic, and health-related factors rather than driven by a single construct. This pattern is consistent with the multifactorial nature of OUD risk and highlights information that may be incompletely represented in routine clinical records.

The LSTM, GRU, and Transformer models generally did not outperform the gradient-boosting approaches despite explicitly modeling event sequences. Much of the useful longitudinal information may already have been captured by summary features such as event frequency, recent values, and average measurements. In the sequential models, laboratory and physical measurement events were represented by their occurrence over time, while their numeric values were summarized across the look-back window as static features. This design may have limited the models' ability to capture changes in these values over time. More broadly, the results suggest that greater model complexity does not necessarily improve prediction when summary features already capture much of the relevant clinical history.

The stronger contribution of survey information at longer observation windows should be interpreted in the context of data availability. Longer look-back periods increased both the opportunity to observe survey responses and the number of retained survey features, which rose from 30 at the 6- and 12-month windows to 108 at 24 months. Thus, the larger gains at 24 months may reflect both accumulation of relevant historical information and greater availability of the survey modality itself. The optimal observation horizon may therefore depend on both the timing of prior information and the likelihood that each data source is available. The strong survey contribution at longer windows also suggests that patient-reported information may provide useful signals for earlier risk assessment and future prevention-oriented screening strategies.

Survey coverage also differed markedly by OUD status, with OUD-negative participants having greater coverage across all three look-back windows. This raises the possibility that missing survey data are informative rather than random. Survey availability may reflect differences in program engagement, enrollment duration, eligibility, completion behavior, or timing relative to the index date. Some of the observed gain may therefore arise from patterns of survey availability in addition to response content. Future ablation analyses that separate availability-related features from substantive survey responses will be important for clarifying these contributions.

Several limitations should be considered. First, OUD status was defined using recorded diagnostic codes and therefore depended on clinical recognition and documentation, which may have led to misclassification of participants with unrecorded OUD. Second, regarding the temporal design, index dates were defined differently for cases and controls, potentially creating differences in observable history, follow-up, and opportunities for survey completion. Third, regarding survey availability, participation was incomplete and differed by outcome status, introducing possible selection effects and informative missingness. Fourth, feature filtering was performed before data splitting, allowing the overall feature distribution to inform feature retention, although OUD status was not used. Given the large cohort, this was unlikely to substantially alter which features were retained. Finally, regarding evaluation, performance was assessed using an internal held-out test set and focused primarily on discrimination. External validation, calibration, subgroup performance, and prospective evaluation are needed before clinical use.

Overall, these findings show both the value and the complexity of integrating patient-reported information with longitudinal EHR data. Survey data provided additional predictive signal, but their contribution depends not only on response content, but also on when and for whom those data are available. Future work should separate survey content from availability effects, assess calibration and fairness, evaluate external generalizability, and determine whether these data improve clinically meaningful OUD screening and risk-assessment workflows.

\section*{Conclusion}

Patient-reported survey information consistently improved OUD prediction beyond
structured EHR data across eight modeling approaches and three look-back
windows. These findings suggest that behavioral, social, functional, and
self-reported health information captures predictive signals that are not fully
represented in routine clinical records. More broadly, integrating
patient-reported data with longitudinal EHRs may strengthen multimodal risk
prediction by providing a more complete view of patient context, while also
highlighting the need to account for differences in survey availability and
completion. With further validation, such approaches may support earlier and
more comprehensive OUD risk assessment and prevention-oriented screening.

\bibliographystyle{vancouver}
\bibliography{refs}

@misc{cdc_oud_dx,
  author      = {{CDC}},
  title       = {Overdose prevention: opioid use disorder: diagnosis},
  type        = {Internet},
  year        = {2024},
  url         = {https://www.cdc.gov/overdose-prevention/hcp/clinical-care/opioid-use-disorder-diagnosis.html},
  lastchecked = {2026 Aug 18},
}

@techreport{nsduh2025,
  author      = {{Substance Abuse and Mental Health Services Administration}},
  title       = {Key substance use and mental health indicators in the United
                 States: results from the 2025 National Survey on Drug Use and
                 Health},
  institution = {Substance Abuse and Mental Health Services Administration},
  address     = {Rockville, MD},
  year        = {2026},
}

@misc{cdc_overdose_2025,
  author      = {{CDC}},
  title       = {NCHS pressroom: U.S. overdose deaths decrease for third
                 consecutive year in 2025},
  type        = {Internet},
  year        = {2026},
  url         = {https://www.cdc.gov/nchs/pressroom/releases/20260513.html},
  lastchecked = {2026 Aug 18},
}

@article{ramirezmedina2025,
  author  = {{Ram{\'i}rez Medina}, C R and {Benitez-Aurioles}, J and
             Jenkins, D A and Jani, M},
  title   = {A systematic review of machine learning applications in predicting
             opioid associated adverse events},
  journal = {npj Digit Med},
  volume  = {8},
  pages   = {30},
  year    = {2025},
  note    = {{d}oi:10.1038/s41746-024-01312-4},
}

@article{Song2024,
  author  = {Song, S L and Dandapani, H G and Estrada, R S and Jones, N W and
             Samuels, E A and Ranney, M L},
  title   = {Predictive models to assess risk of persistent opioid use, opioid
             use disorder, and overdose},
  journal = {J Addict Med},
  volume  = {18},
  pages   = {218--39},
  year    = {2024},
  note    = {{d}oi:10.1097/ADM.0000000000001276},
}

@article{dong2019,
  author  = {Dong, X and Rashidian, S and Wang, Y and Hajagos, J and Zhao, X and
             Rosenthal, R N and others},
  title   = {Machine learning based opioid overdose prediction using electronic
             health records},
  journal = {AMIA Annu Symp Proc},
  volume  = {2019},
  pages   = {389--98},
  year    = {2019},
}

@article{dong2021,
  author  = {Dong, X and Deng, J and Rashidian, S and Abell-Hart, K and
             Hou, W and Rosenthal, R N and others},
  title   = {Identifying risk of opioid use disorder for patients taking opioid
             medications with deep learning},
  journal = {J Am Med Inform Assoc},
  volume  = {28},
  pages   = {1683--93},
  year    = {2021},
  note    = {{d}oi:10.1093/jamia/ocab043},
}

@article{dong2021jbi,
  author  = {Dong, X and Deng, J and Hou, W and Rashidian, S and
             Rosenthal, R N and Saltz, M and others},
  title   = {Predicting opioid overdose risk of patients with opioid
             prescriptions using electronic health records based on temporal
             deep learning},
  journal = {J Biomed Inform},
  volume  = {116},
  pages   = {103725},
  year    = {2021},
  note    = {{d}oi:10.1016/j.jbi.2021.103725},
}

@article{dong2023,
  author  = {Dong, X and Wong, R and Lyu, W and {Abell-Hart}, K and Deng, J and
             Liu, Y and others},
  title   = {An integrated {LSTM}-{HeteroRGNN} model for interpretable opioid
             overdose risk prediction},
  journal = {Artif Intell Med},
  volume  = {135},
  pages   = {102439},
  year    = {2023},
  note    = {{d}oi:10.1016/j.artmed.2022.102439},
}

@article{ding2024,
  author  = {Ding, Z and Dong, X and Liu, Y and Ma, T and Zhao, X and Wong, R and
             others},
  title   = {{HIBERT}: a hybrid clustering {BERT} for interpretable opioid
             overdose risk prediction},
  journal = {AMIA Annu Symp Proc},
  volume  = {2024},
  pages   = {303--12},
  year    = {2024},
}

@techreport{ding2026,
  author      = {Ding, Z and Liu, Y and Ma, T and Wong, R and Leibowitz, G and
                 Littenberg, B and others},
  title       = {A comparative study of feature selection paradigms for
                 structured EHR diagnosis features in OUD prediction},
  institution = {Manuscript under review for the AMIA 2026 Annual Symposium},
  year        = {2026},
  note        = {{a}rXiv:2608.04180. doi:10.48550/ARXIV.2608.04180},
  url         = {https://arxiv.org/abs/2608.04180},
  lastchecked = {2026 Aug 18},
}

@article{Klimas2019,
  author  = {Klimas, J and Gorfinkel, L and Fairbairn, N and Amato, L and
             Ahamad, K and Nolan, S and others},
  title   = {Strategies to identify patient risks of prescription opioid
             addiction when initiating opioids for pain: a systematic review},
  journal = {JAMA Netw Open},
  volume  = {2},
  pages   = {e193365},
  year    = {2019},
  note    = {{d}oi:10.1001/jamanetworkopen.2019.3365},
}

@article{vanRijswijk2019,
  author  = {{van Rijswijk}, S M and {van Beek}, M H C T and Schoof, G M and
             Schene, A H and Steegers, M and Schellekens, A F},
  title   = {Iatrogenic opioid use disorder, chronic pain and psychiatric
             comorbidity: a systematic review},
  journal = {Gen Hosp Psychiatry},
  volume  = {59},
  pages   = {37--50},
  year    = {2019},
  note    = {{d}oi:10.1016/j.genhosppsych.2019.04.008},
}

@article{loeffel2026,
  author  = {Loeffel, L B and Kwak, H R and Shin, J and Lee, M and
             Jester, D J and Dawes, M and others},
  title   = {Associations among social determinants of health and opioid use
             disorder and overdose: an umbrella review},
  journal = {Am J Addict},
  volume  = {35},
  pages   = {339--61},
  year    = {2026},
  note    = {{d}oi:10.1111/ajad.70133},
}

@article{cook2021,
  author  = {Cook, L A and Sachs, J and Weiskopf, N G},
  title   = {The quality of social determinants data in the electronic health
             record: a systematic review},
  journal = {J Am Med Inform Assoc},
  volume  = {29},
  pages   = {187--96},
  year    = {2021},
  note    = {{d}oi:10.1093/jamia/ocab199},
}

@article{ramirez2022,
  author  = {Ramirez, A H and Sulieman, L and Schlueter, D J and
             Halvorson, A and Qian, J and Ratsimbazafy, F and others},
  title   = {The All of Us Research Program: data quality, utility, and
             diversity},
  journal = {Patterns},
  volume  = {3},
  pages   = {100570},
  year    = {2022},
  note    = {{d}oi:10.1016/j.patter.2022.100570},
}

@article{cronin2019,
  author  = {Cronin, R M and Jerome, R N and Mapes, B and Andrade, R and
             Johnston, R and Ayala, J and others},
  title   = {Development of the initial surveys for the All of Us Research
             Program},
  journal = {Epidemiology},
  volume  = {30},
  pages   = {597--608},
  year    = {2019},
  note    = {{d}oi:10.1097/EDE.0000000000001028},
}

@article{giladi2023,
  author  = {Giladi, A M and Shipp, M M and Sanghavi, K K and Zhang, G and
             Gupta, S and Miller, K E and others},
  title   = {Patient-reported data augment prediction models of persistent
             opioid use after elective upper extremity surgery},
  journal = {Plast Reconstr Surg},
  volume  = {152},
  pages   = {358e--66e},
  year    = {2023},
  note    = {{d}oi:10.1097/PRS.0000000000010297},
}

@article{schoenfeld2026,
  author  = {Schoenfeld, A J and Jeyakumar, S and {Morlando Geiger}, J and
             Princic, N and Moynihan, M and Varker, H and others},
  title   = {The incidence of opioid use disorder among people with acute and
             chronic pain managed with prescription opioids in the United
             States: economic and societal burden},
  journal = {J Med Econ},
  volume  = {29},
  pages   = {1355--71},
  year    = {2026},
  note    = {{d}oi:10.1080/13696998.2026.2655086},
}

\end{document}